# B2B Advertising: Joint Dynamic Scoring of Account and Users

Atanu R. Sinha
Adobe Research
atr@adobe.com

Gautam Choudhary
Adobe Research
gautamc@adobe.com

Mansi Agarwal
Adobe Research
mansi10022000@gmail.com

Shivansh Bindal
Adobe Research
shivanshbindal603@gmail.com

Abhishek Pande
Adobe Research
abhishekpande1999@gmail.com

Camille Girabawe
Adobe
girabawe@adobe.com

## ABSTRACT

When a business sells to another business (B2B), the buying business is represented by a group of individuals, termed account, who collectively decide whether to buy. The seller advertises to each individual and interacts with them, mostly by digital means. The sales cycle is long, most often over a few months. There is heterogeneity among individuals belonging to an account in seeking information and hence the seller needs to score the interest of each individual over a long horizon to decide which individuals must be reached and when. Moreover, the buy decision rests with the account and must be scored to project the likelihood of purchase, a decision that is subject to change all the way up to the actual decision, emblematic of group decision making. We score decision of the account and its individuals in a dynamic manner. Dynamic scoring allows opportunity to influence different individual members at different time points over the long horizon. The dataset contains behavior logs of each individual's communication activities with the seller; but, there are no data on consultations among individuals which result in the decision. Using neural network architecture, we propose several ways to aggregate information from individual members' activities, to predict the group's collective decision. Multiple evaluations find strong model performance.

## CCS CONCEPTS

• **Applied computing** → **Electronic commerce**.

## KEYWORDS

group decision, joint scoring, group and individual dynamics

## 1 INTRODUCTION

In business to business marketing (B2B), a buying organization procures products and services from a supplying organization. An example of a product in this setting is an enterprise software system. The buyer is represented by many individuals, to each of whom the supplier advertises and interacts digitally, generating the behavior log data we use. The data also contain the buy/no-buy decision. The individuals belonging to the buyer, termed an *account* (or, *group*), make a collective decision of whether to procure by consulting among themselves, but that data of consultations are not available. B2B's contribution to the Web's success is foremost: "[t]he global B2B eCommerce market valuing USD 12.2 trillion in 2019 is over 6 times that of the B2C [business to consumer] market" [1] , and B2B's digital marketing growth is at par with that of the more commonly studied B2C setting [2]. A B2B setting [1, 15] demands consideration of three research aspects: (a) the group collaboratively decides whether to buy and the decision is dynamic; (b) differences exist among the individuals' interactions with the supplier; (c) over the long purchase cycle, depending on scoring the group dynamics, the supplier engages with the individuals, including allocating costly human salespersons toward some individuals, but not toward all individuals in the group. The scoring models in machine learning (ML) are largely geared toward individual user's action [11–13, 23], but ignore group decision making or group level action, where members of a group collaboratively decide.

These research aspects call for a model that jointly scores (i) prediction of the group decision, (ii) the group dynamics, and (iii) subject to (i) and (ii), the individuals' dynamics. A modeling aspect intrinsic to the problem posed in (i) and (ii) is now noted. The task is not about predicting whether an account procures (hereafter, *converts*) eventually, regardless of the time it takes to convert. Such a task is easier to predict with high accuracy since for every group the ground truth outcome is either buy or no-buy. Instead, the task is to predict purchase probability at every time period so that the supplier can allocate appropriate resources to the group according to the probability. For our model to perform well it must predict no-conversion of the account for prior time periods, and then predict conversion if and when it is likely to occur. Another modeling aspect intrinsic to (iii) is that the group-level dynamic scoring occurs concurrently with scoring every individual in the group for the supplier to target resources differentially to the individuals within a group. All this makes our problem more interesting than previous scoring models.

---

[1] https://www.statista.com/study/44442/statista-report-b2b-e-commerce/
[2] https://cmosurvey.org/wp-content/uploads/2019/08/The_CMO_Survey-Highlights-and_Insights_Report-Aug-2019-1.pdf, pp. 20

Our attention-based network predicts both an account's collective decision to convert [7, 14] and scores each individual in the group, for every time period. The ability of the attention network to assign different decision making weights to different individuals in the group makes it appropriate for the joint scoring task at hand. For group decisions, evaluations on test data of a separate set of accounts, with known decision outcomes, find strong model performance.

Our contributions are in Jointly Scoring:

- Collaborative group decision only from activity data of individual members of the group, without any data of consultations among individual members.
- Dynamics of the group's decision.
- Dynamics of each individual in the group for differently allocating resources to them at different time periods.

## 2 RELEVANT LITERATURE

Individual user-level scoring models dot the ML literature. Individual scoring models include: retargeting users who visit a site but do not buy, early detection of users who are likely to exit without buying by modeling clickstream, early prediction of shopping behavior from live clickstream, and next-best-action marketing and personalization from clickstream [4, 11, 19, 23]. These papers leverage a variety of models from hand-crafted timing and transition models, to traditional classifiers to neural networks. In predicting user behavior for marketing interventions in real-time [13] suggests an ensemble combining neural networks and conventional classifiers. These papers make valuable contributions in the space of individual user behaviors. But, they do not address the joint scoring problem for group decision. A recent work examines outcomes of team movement [18]. In computer science, group decision making is less attended; an exception [17] uses survey data and domain knowledge which provide input for the subjective weights among attributes. Other scoring models are the mainstay of learning to rank research across multiple fields [9, 20], with a long-established literature [16]. The premise of this literature is distinct from that of our group decision premise. For completeness, we make note of the literature on multi-agent learning [8] and multi-armed bandit [6], but it does not examine a group's collective decision.

The literature on group decision making in the social sciences [3, 5, 14] make strong contributions; however, it restricts to reliance on human studies, survey data, or subjective judgment. By contrast, we use ML on actual activity behavior log and jointly score the group and its individuals dynamically.

Our modeling framework comes from Attention Network [2], Hierarchical Attention Network (HAN) in Vision [21], and Natural Language Processing [22]. We modify HAN for group-decision making and joint dynamic scoring. The attention network is suitable for this problem since it can assign different weights to different individuals in the group during the decision making process.

## 3 BENEFITS OF JOINT, DYNAMIC SCORING

Figure 1 shows two samples: for account 229 (left) the account level scores stay flat and at a low value, while individuals' scores have higher values; for account 282 (right) account scores stay flat for several weeks, before shooting up, suggesting a higher likelihood of conversion of account from this point onward. There is no fixed pattern of these dynamics as observed across all accounts in our data; instead, the patterns are idiosyncratic to the behavior of individuals belonging to an account. The group score can be used to rank among different accounts to determine which account receives what kind of attention from the supplier. Conditional on appropriate attention to a group, a supplier can use individual scores to rank-order individuals belonging to the same group, and can decide to send communications accordingly. For example, to some individuals who score high, it can direct salespersons to contact them by phone, while to individuals scoring low it can stay with automated communications (e.g., email). Later, as described in the architecture (Figure 2), the group score is obtained as the output of the network and the individuals' scores are recovered before the aggregation layer.

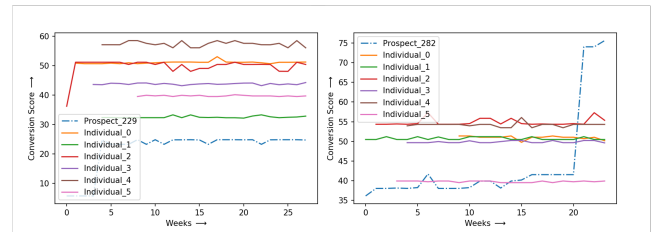

**Figure 1: Scores for two illustrative accounts (groups), 229 on the left and 282 on the right graph.**

## 4 DATA

The data from a supplier comprise time-stamped activities of individuals, who communicate back and forth with the supplier. Our data of logs is confined to activities based on marketing email communications between a supplier and every individual belonging to the accounts, typical in such organizational setting. The data do not show any other forms of communications between the supplier and individuals (e.g., no data on phone or direct email communication with supplier's personnel, nor, any consultations among individuals). Different individuals belong to different accounts, across many industries. An individual is uniquely mapped to an account obeying a many-to-one relationship, both of which are identified through an encrypted ID. Thus, we identify the set of individuals belonging to every account; no individual belongs to multiple accounts. The cardinality of this set varies from 3 to 25, across accounts, reflecting the different sizes of prospective businesses and the size of the probable purchase. Each individual's time-stamped activities, stitched through individual-specific ID, are ordered in time sequence. This sequence has a maximum (minimum) of 114 (1) activities per week per individual with 90%-ile occurring at 8. Some individuals have a short duration of activities (at least a week), while others have activities over many months (at most 35 weeks with 90%-ile at 27). Even for individuals with a prolonged duration of activities, there are periods of no activity. The dataset is not unusual, online suppliers possess this kind of log data.

Description of the data spanning 8 months, is provided in Table 1. The data do not include subject line or body of the email. Individuals who opt-out of emails are not included in the analysis.

The *target label* is a conversion or not for each account, along with the timestamp of when the account converts.

| Types of Information | Features / Variables |
|---|---|
| Dynamic Activities of Individuals | *open email, click email, send email*, unsubscribe email, open sales email, click sales email, send sales email, forwarded email received, forwarded email sent |
| Static Features of Individuals | source of arrival, opt out of email, opt out of phone |
| Static Features of accounts | revenue, number of employees |

**Table 1: Data Description: Data of 9 activities over 8 months for each individual. The top 3 activities are *italicized*. The second row shows 3 categorical static features, vary by individuals. The third row shows group level two static features, they do not vary across individuals within the group; vary by groups.**

Typical of logs of activities in an organizational setting, no data on individuals' consultations within the group are available. There is heterogeneity in time periods at which different individuals, belonging to an account, onboard onto the supplier's system. That is, some individuals have fewer periods of activities and less data for training. The long purchase cycle (property 2) exacerbates this heterogeneity in data for training.

## 5 MODEL ARCHITECTURE FOR SEQUENCE OF ACTIVITIES

The model architecture satisfies the following criteria necessitated by dynamic scoring: (a) Scoring of group decision to convert is obtainable for every time period, where the duration of the time period can be set to any desired value. Criterion (a) calls for defining a time period to organize data. (b) For the same account, (b-i) some individuals establish early contact with the supplier to seek information, while other individuals contact the supplier closer to the decision; and (b-ii) some individuals show no activities in some time periods. Criteria (b-i) and (b-ii) suggest organizing data into rolling windows comprising multiple time periods, and compression of time periods of activities into contiguous periods of activity.

### 5.1 Data Organization

One, for this implementation, the time period is *weekly*, Sunday through Saturday. A reason for the use of a week is that many suppliers score models on a weekly basis. We emphasize that this is not a restriction, other time periods can also be selected. Two, for model training, we seek multiple data samples for each individual belonging to an account, by organizing data into a rolling window of four weeks. The choice of 4 weeks is judgmental to balance the dueling forces of (i) a minimum number of windows to link individuals' activities over weeks, which are used for scoring the account, and (ii) an adequate number of rolling windows from which to sample for training the network. For example, using week 1 to week 4 data for individuals belonging to the account, we predict the probability of account converting in the 5-th week, using week 2 to week 5 data we predict the probability of account converting in the 6-th week, and so on. In unreported results, we use six weeks as well, with little impact on results. Three, based on a histogram of number of activities in a week across all individuals in the data, sequence length of up to 9 activities captures 96 percentile of all data. We choose a vector of fixed length 9 to represent sequence of actions.

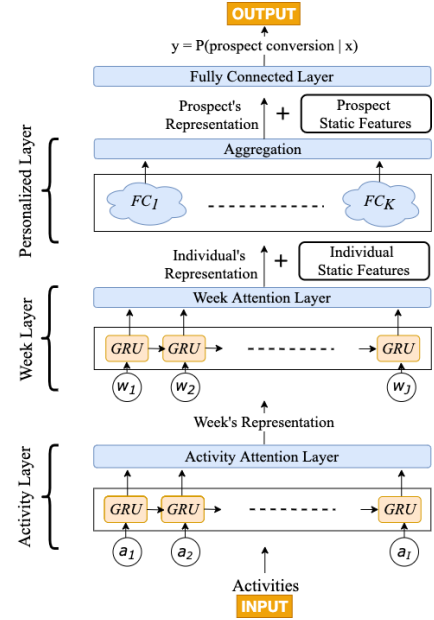

**Figure 2: Network for Sequence of Activities, for an account with multiple individuals. An individual's sequence of activities for a week is encoded in the activity layer, sequence of representations across weeks (rolling window) is encoded in the week layer. Individual-specific representations are encoded in the personalized layer, then aggregated for the group's collective decision to convert.**

### 5.2 Model - Hierarchical Attention Network

In using Hierarchical Attention Network (HAN) [2, 21, 22] our modification arises in the use of aggregation of information across individuals in the third, or, top layer of the network; and in the use of different aggregation functions. See Figure 2. The bottom, activity layer encodes an individual's sequence of activities for one week and outputs that week's representation after passing through an attention layer. We do not predict the following week's sequence of activities, eliminating the need for a decoder. Representations of individual's activities for four weeks, comprising a rolling window, are obtained and encoded in the middle, week layer. After passing through the attention layer, it outputs the individual's representation for a rolling window. In this manner, a representation vector is obtained for each individual, which along with the individual's static features, are passed through the top, fully connected (FC) layer, *specific* to that individual, producing an output for each individual. The outputs from different FC layers for different individuals belonging to the account are combined along with static characteristics of the account. This combination occurs through an aggregate function to give the account's collective decision to convert in the subsequent week. Note that each individual has the same static characteristics since they belong to the same account. The loss is defined as the cross-entropy between the predicted conversion and whether the conversion actually occurred in the subsequent week. We run several experiments with different aggregate functions. The network's technical details follow [22], noting that we have one additional hierarchical level, the third level of personalized layer.

For the activity layer we use a Gated Recurrent Unit (GRU) [10] for encoding sequence of activities and weeks. Given that an account has $K$ individuals, the *activity layer* encodes information of activities performed in a single week by the individual.

The *week layer* encodes information of activities performed by an individual in a rolling window of a specified number of weeks. As explained earlier, given the data at hand, the length of the rolling window is judgmentally set at four weeks.

The *personalized layer* encodes information from weekly activities performed concatenated with static features for $k$-th individual. Static features vary across individuals. This layer computes a value representing that individual's likelihood for conversion.

$$\mathbb{P}_k = \overrightarrow{FC}(o_k), k \in [1, K] \quad (1)$$

Finally, aggregation across individuals is performed. For different aggregation methods, different specifications are used as appropriate and discussed in the next section. After aggregation, static features of the account are concatenated and passed through another fully connected layer to yield a probability score denoting the likelihood of the account to convert. Note that the account static features are common for all individuals belonging to the account.

$$P(\text{account converts}), \mathbb{P} = \overrightarrow{f_{agg}}(\mathbb{P}_1, \mathbb{P}_2, \ldots, \mathbb{P}_K) \quad (2)$$

*5.2.1 Objective Function.* The objective is to minimize cross-entropy loss, L in predicting conversion of the account; that is, $L = -w * y \log(\mathbb{P}) - (1-y) \log(1-\mathbb{P})$, where $\mathbb{P}$ is the predicted probability of conversion for the account, $y$ is a binary target for actual conversion and $w$ is a weight penalty.

## 5.3 Hyperparameters

For the activity layer, the hidden sizes are {40, 56}; for the week layer, it is 20. Weight penalty in loss $w$ lies in {500, 750} to handle the severe class imbalance mentioned above. An Adam optimizer is used for training with a learning rate of 0.01. The number of epochs run is [20,50], with no perceptible change in convergence.

## 6 AGGREGATION METHODS

Several aggregation methods are proposed, classified as: network methods, and statistical functions.

## 6.1 Neural Network Methods

*6.1.1 FNN.* Using the individual's representation vector and account static features we concatenate and pass them through a neural network to give a probability of conversion. We used a vanilla variant with 2 hidden layers followed by a Sigmoid activation to generate the probability of conversion.

*6.1.2 Many-to-One GRU.* Representation vector for each individual is passed through a many-to-one GRU layer in order the individual arrives in supplier's data. This output along with the account's static features is then passed through an FNN with Sigmoid activation to get the probability of conversion.

*6.1.3 Many-to-Many GRU followed by Attention.* Similarly, representation vector for each individual is passed through a many-to-many GRU layer. Its output is then passed through an attention layer to obtain individual-specific contextual weights. This output along with the account's static features is then passed through a simple FNN with Sigmoid activation to get a probability of conversion. The attention mechanism learns weights for each individual's alignment with the group decision and is analogous to the previously described attention mechanisms, but applied to individuals.

## 6.2 Statistical Functions

We implement statistical aggregation of information across individuals by considering the probability of conversion of an account as a function of probabilities of conversion of individuals belonging to the account. Seeking $\mathbb{P}_k$ as a scalar output in the personalized layer, let the probability of conversion of $k$-th individual be, $p_k = \frac{1}{1+exp(-\mathbb{P}_k)}$

*6.2.1 Max probability.* A direct way to decide the conversion of an account is by assuming that if an individual converts, the account converts. Under this assumption we select the maximum probability among all individuals who belong to the account and use it as the probability that the account converts, $\mathbb{P} = \max_{k \in [1,K]} p_k$

*6.2.2 Probability at least one individual converts.* In this variation, we assume that if at least one individual converts, the account converts. Under this assumption we define the probability that account converts as the 1 *minus* the probability none of the individuals converts, $\mathbb{P} = 1 - \prod_{k=1}^{K}(1 - p_k)$

*6.2.3 Geometric Mean.* Since we work with the conversion probability for each individual belonging to the account, another aggregation can be the geometric mean of the probabilities of conversion of individuals, as the probability of conversion of the account, $\mathbb{P} = \sqrt[K]{\prod_{k=1}^{K} p_k}$. For probabilities, an average is better represented by the geometric mean rather than the arithmetic mean.

## 7 EXPERIMENTS AND RESULTS

## 7.1 Data Split

Our test data is a collection of accounts *completely separate* from the set of accounts in the training data. By implication, the set of individuals in the test data is distinct from those in the training data. Hence, there is no information leakage across individuals between train and test datasets. There are a total of 9767 individuals spanning across 2135 accounts in the train set and 1800 individuals spread across 381 accounts in the test set.

## 7.2 Performance Metric

Evaluation is performed for the ability of the proposed and baseline models to predict conversion / no-conversion of each group per time period. The test and train data remain the same across all experiments. Two important empirical implications emerge from the joint dynamic scoring of the group and its individuals. One, within a specific choice of time period and size of the rolling window, we compare multiple proposed models among themselves and with baselines. Thus, the comparisons remain valid since all use the same time period and size of the rolling window. Two, That is, an eventually converted group (positive sample) appearing in the data, appears as non-converted (negative sample) in all time periods prior to the time period in which it converts, rendering very few positive samples overall. Besides, there is no benchmark threshold that can be used to judge the predictive performance of the model for all time periods, since the supplier's threshold for allocating resources to individuals in the group changes over time periods. By contrast, a typical problem in scoring models is whether an account eventually converts, regardless of time. For this static prediction, a benchmark threshold is fixed and come from the proportion of eventual converts (positive samples). Given such benchmark threshold, performance measures such as recall, precision, and f1-score are relevant. In dynamic scoring, the threshold varies based upon the long duration of purchase cycle, and the types of product, and is not discernible from data. The duration can be from 6 months to 18 months. Averaging thresholds over a long duration is not meaningful. Hence, we use AUC to show results, which is threshold independent and can be averaged across time periods.

## 7.3 Baseline Models

We seek baselines that can do joint scoring, to have a like comparison with the proposed models.

*7.3.1 Baseline 1.* It maintains the sequence of activities similar to our proposed model. Scoring of individuals, but who are not subject to a group decision, is the existing standard. Thus, we use individual scoring, with modification for dynamic scoring of individuals and their group. To implement baseline 1, we eliminate the group decision aggregation layer of the architecture (Figure 2) and replace the group level loss function with a loss function for each individual within the group. The target label for each individual in a group is the target label for the group, capturing the idea that when a group converts it is isomorphic to each individual converting. Consistent with our goal, the individual scoring is dynamic. The individuals' scores within a group are added per time period, and normalized to obtain the group dynamic scores.

*7.3.2 Baseline 2.* It uses the frequency of activities, not the sequence of activities. As compared to our proposed model (Figure 2), Baseline 2 maintains the aggregation layer. But, due to the use of frequencies of activities per user, only one attention layer at the level of week, for each individual, is needed; the activity layer is eliminated. Aggregation is done through a Many-to-Many GRU followed by attention. In the data, we have 9 different types of activity, so each week is represented by a vector of length 9. Baseline 2 uses the same rolling windows as in the proposed model.

*7.3.3 Results of Baseline Models.* Table 2 results show that baseline 2, with its frequency of activities, attention layer, and the aggregation layer (AUC = 0.82) outperforms baseline 1, which uses a sequence of activities, attention layer, but without the aggregation layer (AUC = 0.74). This is an indicator of the useful role group level aggregation plays for joint scoring. Baseline 1 conforms to current scoring models for individuals, is devoid of aggregation, and is outperformed.

| Model | AUC |
|---|---|
| Baseline 1: Activity Sequence, Individual loss, No Aggregation | 0.74 |
| Baseline 2: Activity Frequency, Group loss, Aggregation | 0.82 |

**Table 2: Baseline Model Performance: Frequency of Activities with Aggregation performs appreciably better.**

## 7.4 Our Models: Experiments 1-6

First, we run experiments for the architecture with the sequence of actions as input and for multiple types of aggregation described above. We run a set of six experiments for this architecture, the results of which are shown in Table 3. Experiments 1-3 in Table 3 with aggregation using FNN and GRUs have AUC of 0.86-0.87, which are higher than that of experiments 4-6 with statistical aggregation functions, with AUC lying between 0.67 to 0.83. Among statistical functions, the 'Maximum probability' function of experiment 4 performs much better with AUC 0.83. One intuitive reason is that the functions 'Probability at least one individual converts' (experiment 5) and 'Geometric Mean' (experiment 6) are an aggregation of probabilities, which do not recognize the differences in influence and engagement among individuals. The function 'Maximum probability' recognizes the individual with the highest influence and engagement with the supplier.

| Experiments 1-6 with different aggregation methods | AUC |
|---|---|
| **Neural Network methods** | |
| 1. FNN Layer | 0.86 |
| 2. Many-to-One GRU | 0.87 |
| 3. Many-to-Many GRU and Attention Layer | 0.87 |
| **Statistical functions** | |
| 4. Maximum probability converts | 0.83 |
| 5. Probability at least one individual converts | 0.67 |
| 6. Geometric Mean | 0.69 |

**Table 3: Model Performance: Sequence of Activities. All proposed models 1-6 appreciably outperform the baseline-sequence model. Neural network aggregation methods perform better than use of statistical functions for aggregation.**

## 7.5 Our Models, time-LSTMs: Experiments 7-9

This set of experiments 7-9 continue to utilize the architecture with the sequence of actions as input, but with a difference from the previous experiments 1-6. For dynamic modeling of individuals' behaviors, the time elapsed between successive actions can potentially signal useful information about their behaviors [24]. In the bottom, activity layer of the model architecture shown in Figure 2, we concatenate *time between successive actions* to the sequence of actions. In experiments 7-9, we use the three best performing aggregation methods from Table 3, which correspond to Experiments 1-3. The results are shown in Table 4. Comparing results of Experiments 7-9 with those of Experiments 1-3, we find a small decrease of AUC, namely, between .01 and .02, across three experiments. This suggests that the inclusion of time elapsed between successive activities is not contributing to an improvement in AUC scores as compared to results in experiments 1-3.

| Experiments 7-9 with time-LSTMs | AUC |
|---|---|
| 7. FNN Layer | 0.85 |
| 8. Many-to-One GRU | 0.85 |
| 9. Many-to-Many GRU and Attention Layer | 0.86 |

Table 4: Model Performance: Sequence of Activities, time-LSTMs. All proposed neural network models, outperform Baseline1 strongly, and Baseline 2 as well.

## 7.6 Insights from Model Comparisons

Extensive experiments run score both group and individuals in the group. Several insights emerge: (1) Use of neural network aggregation functions which then feed back a group level loss, makes an appreciable difference to the performance, relative to the use of individual level loss functions (without aggregation). This is evident by much higher AUCs, 0.86 - 0.87, of experiments 1-3 (Table 3) with Baseline 1 AUC of 0.74 (Table 2). (2) The use of statistical aggregation (Table 3) shows a mixed performance relative to no-aggregation (Table 2) since maximum probability's AUC of 0.83 is higher but than of the other two (experiments 5 and 6, AUC 0.67, 0.69) are lower than baseline 1 AUC of 0.74. (3) Using frequency of activities, instead of sequence of activities, although maintaining aggregated group level loss, yields a good Baseline 2 AUC of 0.82 (Table 2); however, the sequence of activities with an identical set up of Many-to-Many GRUs gives a significant jump in AUC to 0.87, as in experiment 3 of Table 3. The sequence of activities does inform the model more effectively than the frequency of activities. (4) Addition of time between successive actions as input over the above sequence does not provide any jump in performance as observed comparing its AUCs, 0.85 - 0.86, (Table 4) to that of AUCs 0.86-0.87 in experiments 1-3 (Table 3). For deriving additional insight, we study the Pearson correlation between the weekly sum of each group's total activities and the weekly model based group score. A regression analysis finds no difference (p-value = 0.86), between the converted versus not converted groups. Thus, the volume of activities by individuals in a group is not a good indicator of the tendency for the group to convert.

## 8 CONCLUSION

Toward joint scoring models in ML, we predict a group and its individuals dynamically, subject to the group decision. We extend an attention network to assign differential decision making weights, akin to differential influences, to different individuals in the group. Two classes of aggregation methods are proposed - network based and statistical function based - and within each class, multiple methods are compared. Performance of several experiments for predicting the dynamics of the decision, relative to respective baselines, finds strong support for our approach for joint scoring of the group decision dynamically. The dynamic scores predict the time interval of conversion by examining when the score crosses a desired threshold, resulting in differential resource allocation among accounts, and among individuals within each account. We hope future ML research focuses more on group decision making.

## REFERENCES
[1] Yuval Atsmon. 2016. How nimble resource allocation can double your company's value. https://www.mckinsey.com/business-functions/strategy-and-corporate-finance/our-insights/how-nimble-resource-allocation-can-double-your-companys-value.
[2] Dzmitry Bahdanau, Kyunghyun Cho, and Yoshua Bengio. 2014. Neural machine translation by jointly learning to align and translate. *arXiv preprint arXiv:1409.0473* (2014).
[3] Gary S Becker. 1974. A theory of social interactions. *Journal of political economy* 82, 6 (1974), 1063–1093.
[4] Luca Bigon, Giovanni Cassani, Ciro Greco, Lucas Lacasa, Mattia Pavoni, Andrea Polonioli, and Jacopo Tagliabue. 2019. Prediction is very hard, especially about conversion. Predicting user purchases from clickstream data in fashion e-commerce. *arXiv preprint arXiv:1907.00400* (2019).
[5] Duncan Black. 1948. On the rationale of group decision-making. *Journal of political economy* 56, 1 (1948), 23–34.
[6] Djallel Bouneffouf, Irina Rish, and Charu Aggarwal. 2020. Survey on applications of multi-armed and contextual bandits. In *2020 IEEE Congress on Evolutionary Computation (CEC)*. IEEE, 1–8.
[7] Felix C Brodbeck, Rudolf Kerschreiter, Andreas Mojzisch, and Stefan Schulz-Hardt. 2007. Group decision making under conditions of distributed knowledge: The information asymmetries model. *Academy of Management Review* 32, 2 (2007), 459–479.
[8] Lucian Busoniu, Robert Babuska, and Bart De Schutter. 2008. A comprehensive survey of multiagent reinforcement learning. *IEEE Transactions on Systems, Man, and Cybernetics, Part C (Applications and Reviews)* 38, 2 (2008), 156–172.
[9] Fatih Cakir, Kun He, Xide Xia, Brian Kulis, and Stan Sclaroff. 2019. Deep metric learning to rank. In *Proceedings of the IEEE/CVF Conference on Computer Vision and Pattern Recognition*. 1861–1870.
[10] Kyunghyun Cho, Bart Van Merriënboer, Dzmitry Bahdanau, and Yoshua Bengio. 2014. On the properties of neural machine translation: Encoder-decoder approaches. *arXiv preprint arXiv:1409.1259* (2014).
[11] Tobias Hatt and Stefan Feuerriegel. 2020. Early detection of user exits from clickstream data: A markov modulated marked point process model. In *Proceedings of The Web Conference 2020*. 1671–1681.
[12] Wael Jabr, Abhijeet Ghoshal, Yichen Cheng, and Paul A Pavlou. 2020. Maximizing Revisiting and Purchasing: A Clickstream-Based Approach to Enhance Individual-Level Customer Conversion. *Available at SSRN 3665399* (2020).
[13] Dennis Koehn, Stefan Lessmann, and Markus Schaal. 2020. Predicting online shopping behaviour from clickstream data using deep learning. *Expert Systems with Applications* 150 (2020), 113342.
[14] Tamar Kugler, Edgar E Kausel, and Martin G Kocher. 2012. Are groups more rational than individuals? A review of interactive decision making in groups. *Wiley Interdisciplinary Reviews: Cognitive Science* 3, 4 (2012), 471–482.
[15] Gary L Lilien. 2016. The B2B knowledge gap. *International Journal of Research in Marketing* 33, 3 (2016), 543–556.
[16] Tie-Yan Liu. 2011. Learning to rank for information retrieval. (2011).
[17] Hsu-Shih Shih, Huan-Jyh Shyur, and E Stanley Lee. 2007. An extension of TOPSIS for group decision making. *Mathematical and computer modelling* 45, 7-8 (2007), 801–813.
[18] Omid Shokrollahi, Bahman Rohani, and Amin Nobakhti. 2021. Predicting the outcome of team movements – Player time series analysis using fuzzy and deep methods for representation learning. arXiv:2109.07570 [cs.LG]
[19] Arthur Toth, Louis Tan, Giuseppe Di Fabbrizio, and Ankur Datta. 2017. Predicting shopping behavior with mixture of RNNs. In *eCOM@ SIGIR*.
[20] Xuanhui Wang, Nadav Golbandi, Michael Bendersky, Donald Metzler, and Marc Najork. 2018. Position bias estimation for unbiased learning to rank in personal search. In *Proceedings of the Eleventh ACM International Conference on Web Search and Data Mining*. 610–618.
[21] Yilin Wang, Suhang Wang, Jiliang Tang, Neil O'Hare, Yi Chang, and Baoxin Li. 2016. Hierarchical attention network for action recognition in videos. *arXiv preprint arXiv:1607.06416* (2016).
[22] Zichao Yang, Diyi Yang, Chris Dyer, Xiaodong He, Alex Smola, and Eduard Hovy. 2016. Hierarchical attention networks for document classification. In *Proceedings of the 2016 conference of the North American chapter of the association for computational linguistics: human language technologies*. 1480–1489.
[23] Jinyoung Yeo, Seung-won Hwang, Eunyee Koh, Nedim Lipka, et al. 2018. Conversion prediction from clickstream: Modeling market prediction and customer predictability. *IEEE Transactions on Knowledge and Data Engineering* 32, 2 (2018), 246–259.
[24] Yu Zhu, Hao Li, Yikang Liao, Beidou Wang, Ziyu Guan, Haifeng Liu, and Deng Cai. 2017. What to Do Next: Modeling User Behaviors by Time-LSTM.. In *IJCAI*, Vol. 17. 3602–3608.